%% file: main.tex
\documentclass[10pt,twocolumn,letterpaper]{article}

\usepackage{cvpr}             
\usepackage{svg}


\definecolor{cvprblue}{rgb}{0.21,0.49,0.74}
\usepackage[pagebackref,breaklinks,colorlinks,citecolor=cvprblue]{hyperref}

\def\paperID{0} 
\def\confName{CVPR}
\def\confYear{2025}

\title{%
DIVE: Deep-search Iterative Video Exploration \\
A Technical Report for the CVRR Challenge at CVPR 2025
}

\author{Umihiro Kamoto \quad\quad Tatsuya Ishibashi \quad\quad Noriyuki Kugo\\
Panasonic Connect Co., Ltd.\\
{\tt\small \{kamoto.umihiro, ishibashi.tatsuya001, kugou.noriyuki\}@jp.panasonic.com}
}

\begin{document}
\maketitle
\input{sec/0_abstract}    
\input{sec/1_intro}
\input{sec/2_approach}
\input{sec/3_experiments}
\input{sec/4_conclusion}
{
    \small
    \bibliographystyle{ieeenat_fullname}
    \bibliography{main}
}


\end{document}

%% file: sec/0_abstract.tex
\begin{abstract}
In this report, we present the winning solution that achieved the 1st place in the Complex Video Reasoning \& Robustness Evaluation Challenge 2025.
This challenge evaluates the ability to generate accurate natural language answers to questions about diverse, real-world video clips. It uses the Complex Video Reasoning and Robustness Evaluation Suite (CVRR-ES) benchmark, which consists of 214 unique videos and 2,400 question-answer pairs spanning 11 categories.
Our method, \textbf{DIVE (Deep-search Iterative Video Exploration)}, adopts an iterative reasoning approach, in which each input question is semantically decomposed and solved through stepwise reasoning and progressive inference. This enables our system to provide highly accurate and contextually appropriate answers to even the most complex queries.
Applied to the CVRR-ES benchmark, our approach achieves 81.44\% accuracy on the test set, securing the top position among all participants.
This report details our methodology and provides a comprehensive analysis of the experimental results, demonstrating the effectiveness of our iterative reasoning framework in achieving robust video question answering.
The code is available at \url{https://github.com/PanasonicConnect/DIVE}.
\end{abstract}

%% file: sec/1_intro.tex
\section{Introduction}
\label{sec:intro}

Video understanding is a critical research challenge in both computer vision and natural language processing. In recent years, the Video Question Answering (VQA) task, which requires models to answer a question about video clips, has emerged as a comprehensive benchmark for assessing video understanding capabilities. 
Approaches to VQA have become increasingly diverse, with extensive research on end-to-end methods~\cite{Ren2023TimeChat, li2024llamavid, Wang2022InternVideoGV, li2023blip, wang2023videollava, Yang2022ZeroShotVQ} and approaches that convert video content into image captions and leverage large language models (LLMs) to generate answers~\cite{Wang2024VideoTreeAT, Maaz2023VideoChatGPT,  Zhang2023ASL, Wang2023LifelongMemoryLL, Zhang2024HCQAE}. More recently, methods that incorporate AI agents have also gained attention. For example, some approaches dynamically generate specialized agents based on the question, while others employ agents to select important video frames~\cite{Fan2024VideoAgentAM, wang2024videoagent, kugo2024vdma, Kugo2025VideoMultiAgentsAM, Zhi2025VideoAgent2ET, Shang2024TraveLERAM}.
Iterative reasoning approaches, such as OpenAI’s DeepResearch~\cite{openai2025deepresearch}, which employ multi-step inference strategies, have advanced LLMs by enabling a deeper and more robust understanding of complex questions.
However, these techniques are still rarely applied to VQA, especially for video understanding. As a result, many existing VQA methods answer questions in a straightforward manner, often overlooking the underlying intent and producing less contextually appropriate responses. To address these gaps, we propose incorporating intent estimation and iterative reasoning to further improve video question answering.


In this paper, we present DIVE (Deep-search Iterative Video Exploration), a novel framework that combines semantic decomposition, intent estimation, and iterative inference to generate accurate answers to questions about given video clips. Specifically, DIVE breaks down each question into sub-questions based on the underlying intent, and solves them through an iterative process to generate contextually appropriate and precise answers.
Additionally, we propose an object-centric video summarization method. This method uses object detection techniques to create summaries focused on the appearance and spatio-temporal transitions of key objects in each scene. This leads to improved overall performance in video understanding.
The main contributions of this work are as follows:

\begin{itemize}
    \item We propose DIVE, a new framework that answers complex video questions by breaking them down into sub-questions and solving them through an iterative reasoning process.
    \item We incorporate intent estimation into the VQA task to enable answers that better capture the underlying intent behind each question.
    \item We develop a novel object-centric video summarization method that enhances video comprehension by capturing key object transitions.
\end{itemize}

Our results demonstrate that our method effectively and robustly addresses the complex queries featured in the CVRR-ES benchmark.


%% file: sec/2_approach.tex
\section{Methodology}
\label{sec:formatting}

\hyphenpenalty=10000
\exhyphenpenalty=10000

\begin{figure}[t]
  \centering
  \includegraphics[width=0.70\linewidth]{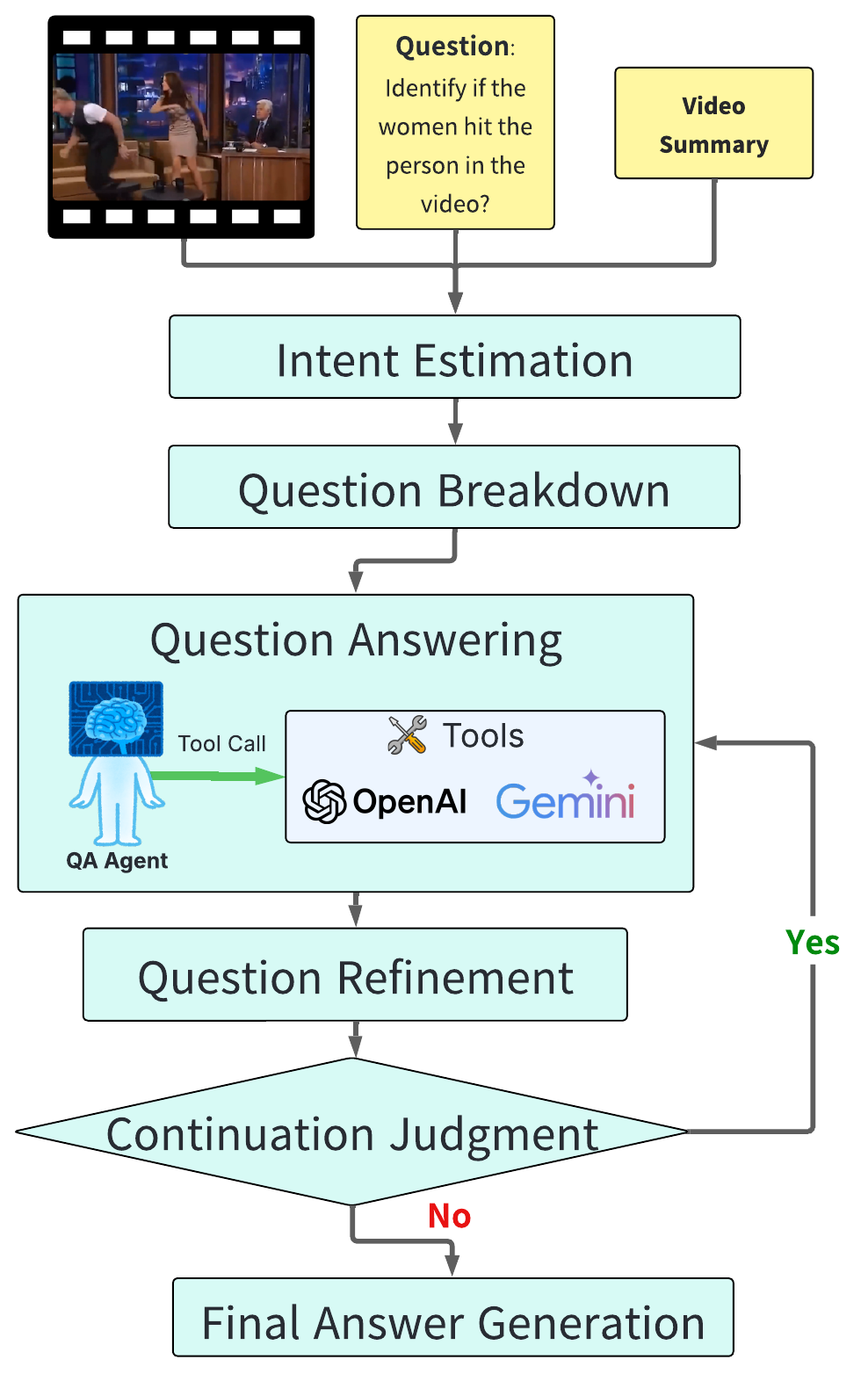}
  \captionsetup{justification=raggedright, singlelinecheck=false}
  \caption{Overall architecture of DIVE: Achieving robust and accurate VideoQA by iteratively solving sub-questions decomposed by the question breakdown module.}
  \label{fig:architecture}
\end{figure}

Figure~\ref{fig:architecture} presents the overall architecture of our system, which incrementally and iteratively enhances video understanding and answer accuracy through a six-step process: (1) intent estimation, (2) question breakdown into sub-questions, (3) agent-based answering of sub-questions, (4) refinement of remaining sub-questions, (5) loop continuation judgment, and (6) final answer generation by integrating all answers to sub-questions. The following subsections describe each step in detail.

\subsection{Step 1: Intent Estimation}

In Step 1, intent estimation is performed on the input question. The system interprets what the question is fundamentally asking by considering both the question text and the video context. Specifically, a video summary is used as the key visual information from the video. The question and this video summary are then combined to estimate and describe in detail the underlying intent of the question.

\subsection{Step 2: Question Breakdown}

In Step 2, the question is decomposed into multiple sub-questions. At this step, both the question text and the video summary information are provided as input, and relevant sub-questions are generated based on the original question. This decomposition enables the system to focus on specific parts or aspects of the video.
Furthermore, information about the video analysis tools to be used by the agent in Step 3 is also considered when generating the sub-questions. By tailoring the sub-questions to match the analytical capabilities of the subsequent agent, the system enables more effective and targeted video analysis.

\subsection{Step 3: Question Answering}

In Step 3, an agent is employed to answer the sub-question with the highest priority in each loop. The remaining sub-questions are retained for further refinement in the next step. For each sub-question, the agent utilizes the following dedicated video analysis tools to produce appropriate responses:
\begin{itemize}
    \item \textbf{Gemini 2.5 Pro Tool}: Analyzes one frame per second along with the corresponding audio. This tool excels in temporal reasoning, recognizing audio cues. Because it processes the entire video at a consistent temporal resolution, it is particularly useful for achieving a comprehensive understanding of the overall video content.
    \item \textbf{GPT-4.1 Tool}: Based on the sub-question and video summary information, this tool identifies key temporal segments of interest, samples frames from those segments, and performs detailed analysis using GPT-4.1. By selecting and analyzing 8-16 frames that are most relevant to the sub-question, this tool enables detailed analysis of specific temporal segments within the video.
\end{itemize}
By leveraging these tools, the agent can generate optimal answers for each sub-question.

\subsection{Step 4: Question Refinement}
In Step 4, the remaining sub-questions are refined and adjusted as needed based on the agent’s responses. For example, if a sub-question regarding the existence of an object receives a negative answer, additional follow-up sub-questions may be generated to reconfirm the result. In this way, the sub-questions are flexibly modified or supplemented according to the situation, enabling the confirmation of necessary information and more precise video understanding through optimized question flow and content.

\subsection{Step 5: Continuation Judgment}
In Step 5, the system decides whether to continue the loop. The need for additional information is assessed based on the agent's responses and the current set of sub-questions. If additional information is required to confidently answer the original question, the process returns to Step 3. In contrast, when sufficient information has been gathered, particularly for straightforward questions, the system terminates the loop to avoid ``overthinking'' and maintain the quality of the answers. This iterative procedure enables continuous and adaptive information gathering, facilitating a deeper understanding of the video while avoiding unnecessary analysis.

\subsection{Step 6: Final Answer Generation}
In Step 6, the final answer is generated. The system synthesizes the agent’s responses to the sub-questions together with the video summary to produce a comprehensive answer to the original question. By integrating these sources of information, the system provides a contextually accurate and precise response.

\subsection{Video Summarization Using Object Detection}
Our video summarization process leverages object detection results to generate context-aware summaries for each video. The procedure consists of three main steps:

\begin{enumerate}
    \item \textbf{Object Label Extraction:} 32 evenly spaced frames are sampled from each video, and object labels are extracted from these frames using GPT-4.1.
    \item \textbf{Object Detection:} Using the extracted object labels as prompts, we perform object detection on all video frames using Grounding DINO~\cite{liu2024grounding}, capturing the presence and spatial locations of these objects throughout the video.
    \item \textbf{Video Summary Generation:} A video summary is generated using GPT-4.1 by integrating the sampled frames and the object detection results, describing key objects, their transitions, and interactions throughout the video.
\end{enumerate}

This approach enables the creation of video summaries that reflect both global context and fine-grained object-level information, thereby supporting more accurate video understanding and downstream tasks.

%% file: sec/3_experiments.tex
\section{Experiments}

In this section, we present the evaluation of our proposed DIVE on the CVRR-ES benchmark. 
We first provide implementation details of our DIVE architecture in Section~\ref{subsec:implementation}.
In Section~\ref{subsec:results}, we compare our method with existing approaches.
We then conduct an in-depth ablation study in Section~\ref{subsec:ablation}, where we analyze the impact of each component and include detailed case studies to illustrate the effectiveness and characteristics of our approach.

\subsection{Implementation Details}
\label{subsec:implementation}
Our proposed method, DIVE, was implemented using the open-source LangGraph library~\footnote{LangGraph, \url{https://www.langchain.com/langgraph}}. For all steps, we used an agent and LLM based on OpenAI's GPT-4.1~\footnote{OpenAI, \url{https://openai.com}} (version gpt-4.1-2025-04-14).
For the tools invoked by the agent during Step 3 (Question Answering), we utilized Gemini 2.5 Pro, provided by Google Vertex AI~\footnote{Vertex AI, \url{https://cloud.google.com/vertex-ai}} (version gemini-2.5-pro-preview-05-06), as the Gemini 2.5 Pro Tool for video analysis. For frame-level image analysis, we employed OpenAI's GPT-4.1 within the GPT-4.1 Tool.
The temperature parameter for the LLMs used in each tool was set to 1.0, and to 0.0 for the agent itself.
Additionally, the maximum number of reasoning steps in DIVE was set to 25.

\begin{table}[t]
    \centering
    \caption{Comparison with EvalAI leaderboard entries and paper-reported baselines on the CVRR-ES validation set.}
    \begin{tabular}{cc}
        \hline
        Method & Acc. (\%) \\
        \hline
        Baseline (GPT4V)~\cite{Khattak2024cvrres} & 70.78 \\
        Baseline (GPT-4o)~\cite{Khattak2024cvrres} & 75.03 \\
        \hline
        FRI & 53 \\
        Host\_6403\_Team & 63 \\
        NJUST\_\_KMG & 85 \\
        PCIEgogogo & 88 \\
        \hline
        DIVE & \textbf{91.55} \\
        \hline
    \end{tabular}
    \label{tab:result1}
\end{table}

\begin{table}[t]
    \centering
    \caption{Comparison with the leaderboard entries announced by the organizing committee on the CVRR-ES test set.}
    \begin{tabular}{cc}
        \hline
        Method & Acc. (\%) \\
        \hline
        aaa\_vlm & 65.23 \\
        PCIEgo & 73.69 \\
        PCIEgogogo & 74.95 \\
        PCIE & 75.32 \\
        love\_liang & 77.84 \\
        NJUST\_\_KMG & 78.02\\
        \hline
        DIVE & \textbf{81.44} \\
        \hline
    \end{tabular}
    \label{tab:result2}
\end{table}


\subsection{Main Results}
\label{subsec:results}

Table~\ref{tab:result1} summarizes the performance of existing methods and participants on EvalAI public leaderboard for the CVRR-ES validation set, while Table~\ref{tab:result2} presents the results for the test set from the leaderboard announced by the organizing committee.
As shown in Table~\ref{tab:result1}, our proposed method achieves an accuracy of 91.55\% on the validation set, significantly surpassing both the highest previously reported score of 75.03\% from the CVRR-ES paper~\cite{Khattak2024cvrres} and the scores of all other teams on the public leaderboard.
Likewise, as shown in Table~\ref{tab:result2}, our method achieves the best performance on the test set with an accuracy of 81.44\%, again outperforming all competing approaches on the leaderboard.
These results demonstrate the superior robustness and accuracy of our approach on the CVRR-ES benchmark compared to existing methods.

\begin{table}[t]
\centering
\setlength{\tabcolsep}{5pt}
\small
\caption{Ablation study on CVRR-ES validation set. GPT-4.1 refers to direct API use; QA Agent employs both GPT-4.1 and Gemini 2.5 Pro tools. Checkmarks (\checkmark) indicate enabled components, and values in parentheses show incremental accuracy gains with each added component.}
\begin{tabular}{@{}l|ccc|c@{}}
\toprule
QA Method   & Breakdown  & Intent     & Summary    & Acc. (\%) \\
\midrule
GPT-4.1     &            &            &            & 81.49 \phantom{(+0.00)} \\
GPT-4.1     & \checkmark &            &            & 85.83 (+4.34) \\
QA Agent    & \checkmark & \checkmark &            & 87.58 (+1.75) \\
QA Agent    & \checkmark & \checkmark & \checkmark & 88.00 (+0.42) \\
\bottomrule
\end{tabular}
\label{tab:ablation}
\end{table}

\subsection{Ablation Study}
\label{subsec:ablation}

We conduct an ablation study to evaluate the effectiveness of key components in our proposed method. We progressively integrate individual modules into a base pipeline and assess the contribution of each component to overall performance. Specifically, we investigate three key components: agent-based QA for answering sub-questions, intent estimation in Step 1, and video summarization using object detection for Step 2 (Question Breakdown). All experiments are conducted on the CVRR-ES validation set following the evaluation protocol of~\cite{Khattak2024cvrres}.

Table~\ref{tab:ablation} shows the results of our ablation study on the CVRR-ES validation set. Starting from the GPT-4.1 baseline, adding the Question Breakdown module increases accuracy from 81.49\% to 85.83\%. Incorporating Intent Estimation and QA Agent further boosts performance to 87.58\%, and the addition of Video Summarization yields the best result of 88.00\%. These findings highlight the incremental contribution of each component to the overall effectiveness of our approach.

\begin{figure}[t]
  \centering
  \includegraphics[width=0.95\linewidth]{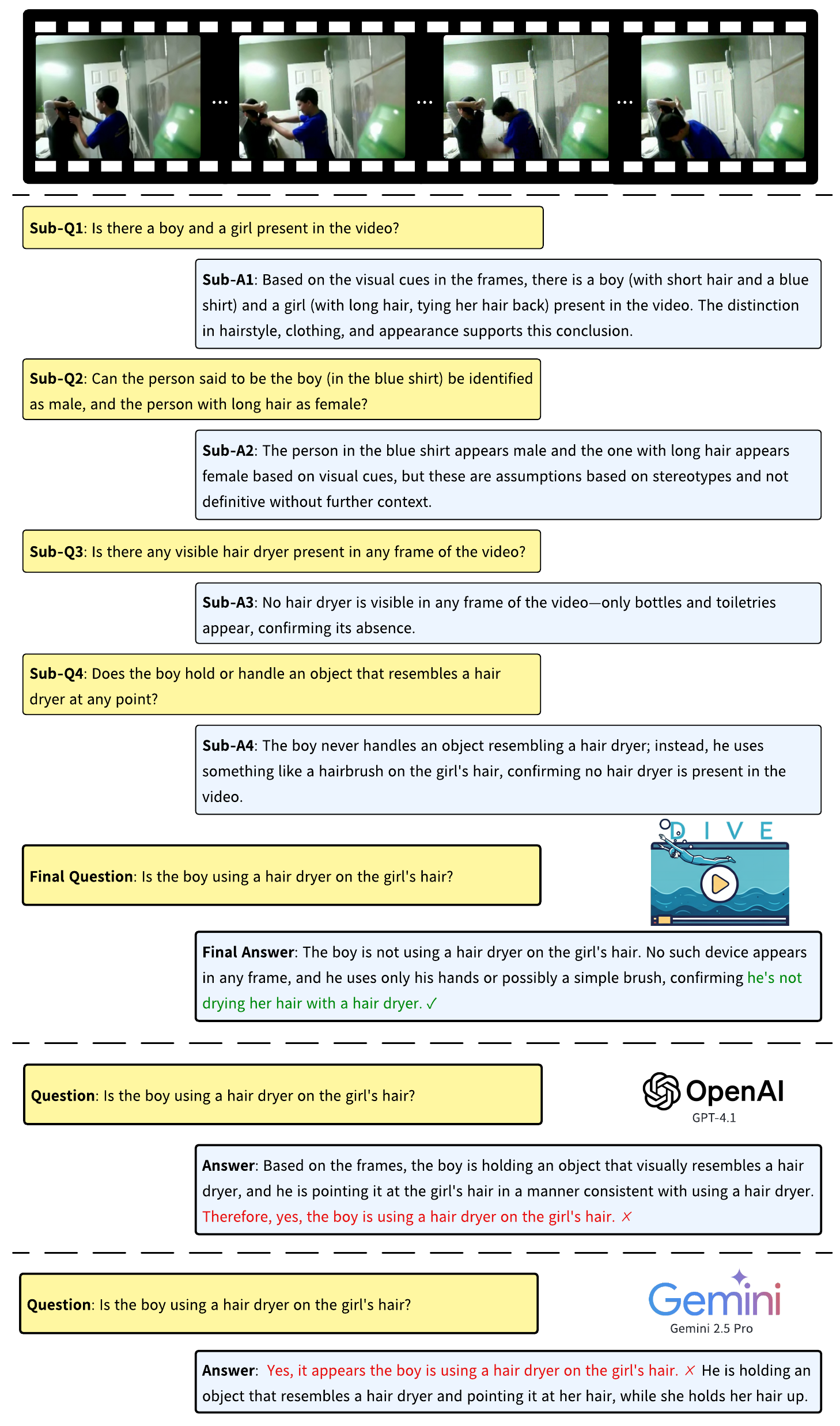}
  \captionsetup{justification=raggedright, singlelinecheck=false}
  \caption{An example result from the CVRR-ES benchmark where Gemini 2.5 Pro and GPT-4.1 fail, but DIVE achieves the correct answer by breaking down the question and solving it iteratively.}
  \label{fig:example}
\end{figure}

In addition to quantitative results, Figure~\ref{fig:example} presents a representative example from the CVRR-ES benchmark where both Gemini 2.5 Pro and GPT-4.1 fail to produce the correct answer. In contrast, our method, DIVE, successfully solves the problem by breaking down the complex question into sub-questions and applying iterative reasoning. This qualitative result further underscores the strength of our modular approach, particularly the question breakdown and iterative reasoning components, in addressing complex queries that require deeper understanding and iterative inference.

%% file: sec/4_conclusion.tex
\section{Conclusion}

In this report, we have proposed and evaluated a robust video question answering method that achieved 1st place in the Complex Video Reasoning \& Robustness Evaluation Challenge 2025.
Our approach combines iterative reasoning, intent estimation, and video summarization using object detection, enabling highly accurate and contextually appropriate answers to complex questions in the CVRR-ES dataset.
Experimental results show that our method achieves accuracies of 91.55\% on the validation set and 81.44\% on the test set, significantly outperforming existing methods and other participating teams.
Furthermore, ablation studies confirmed the significant contribution of each module, such as Question Breakdown, Intent Estimation, and Video Summarization, to overall performance.

Future work will focus on further optimizing computational efficiency, exploring dynamic module selection, and evaluating the generalizability of our approach to other video QA benchmarks.

%% file: main.bbl
\begin{thebibliography}{20}
\providecommand{\natexlab}[1]{#1}
\providecommand{\url}[1]{\texttt{#1}}
\expandafter\ifx\csname urlstyle\endcsname\relax
  \providecommand{\doi}[1]{doi: #1}\else
  \providecommand{\doi}{doi: \begingroup \urlstyle{rm}\Url}\fi

\bibitem[Fan et~al.(2024)Fan, Ma, Wu, Du, Li, Gao, and Li]{Fan2024VideoAgentAM}
Yue Fan, Xiaojian Ma, Rujie Wu, Yuntao Du, Jiaqi Li, Zhi Gao, and Qing Li.
\newblock Videoagent: A memory-augmented multimodal agent for video understanding.
\newblock \emph{ArXiv}, abs/2403.11481, 2024.

\bibitem[khattak et~al.(2024)khattak, Naeem, Hassan, Muzzamal, Tombari, Khan, and Khan]{Khattak2024cvrres}
Muhammad~Uzair khattak, Muhammad~Ferjad Naeem, Jameel Hassan, Naseer Muzzamal, Federcio Tombari, Fahad~Shahbaz Khan, and Salman Khan.
\newblock How good is my video lmm? complex video reasoning and robustness evaluation suite for video-lmms.
\newblock \emph{arXiv:2405.03690}, 2024.

\bibitem[Kugo et~al.(2024)Kugo, Ishibashi, Ono, and Sato]{kugo2024vdma}
Noriyuki Kugo, Tatsuya Ishibashi, Kosuke Ono, and Yuji Sato.
\newblock Vdma: Video question answering with dynamically generated multi-agents.
\newblock \emph{ArXiv}, abs/2407.03610, 2024.

\bibitem[Kugo et~al.(2025)Kugo, Li, Li, Gupta, Khatua, Jain, Patel, Kyuragi, Ishii, Tanabiki, Kozuka, and Adeli]{Kugo2025VideoMultiAgentsAM}
Noriyuki Kugo, Xiang Li, Zixin Li, Ashish Gupta, Arpandeep Khatua, Nidhish Jain, Chaitanya Patel, Yuta Kyuragi, Yasunori Ishii, Masamoto Tanabiki, Kazuki Kozuka, and Ehsan Adeli.
\newblock Videomultiagents: A multi-agent framework for video question answering.
\newblock \emph{ArXiv}, abs/2504.20091, 2025.

\bibitem[Li et~al.(2023)Li, Li, Savarese, and Hoi]{li2023blip}
Junnan Li, Dongxu Li, Silvio Savarese, and Steven C~H Hoi.
\newblock Blip-2: Bootstrapping language-image pre-training with frozen image encoders and large language models.
\newblock \emph{ICML}, 2023.

\bibitem[Li et~al.(2024)Li, Wang, and Jia]{li2024llamavid}
Yanwei Li, Chengyao Wang, and Jiaya Jia.
\newblock Llama-vid: An image is worth 2 tokens in large language models.
\newblock 2024.

\bibitem[Lin et~al.(2023)Lin, Zhu, Ye, Ning, Jin, and Yuan]{wang2023videollava}
Bin Lin, Bin Zhu, Yang Ye, Munan Ning, Peng Jin, and Li Yuan.
\newblock Video-llava: Learning united visual representation by alignment before projection.
\newblock In \emph{Conference on Empirical Methods in Natural Language Processing}, 2023.

\bibitem[Liu et~al.(2024)Liu, Zeng, Ren, Li, Zhang, Yang, Jiang, Li, Yang, Su, et~al.]{liu2024grounding}
Shilong Liu, Zhaoyang Zeng, Tianhe Ren, Feng Li, Hao Zhang, Jie Yang, Qing Jiang, Chunyuan Li, Jianwei Yang, Hang Su, et~al.
\newblock Grounding dino: Marrying dino with grounded pre-training for open-set object detection.
\newblock In \emph{European Conference on Computer Vision}, pages 38--55. Springer, 2024.

\bibitem[Maaz et~al.(2024)Maaz, Rasheed, Khan, and Khan]{Maaz2023VideoChatGPT}
Muhammad Maaz, Hanoona Rasheed, Salman Khan, and Fahad~Shahbaz Khan.
\newblock Video-chatgpt: Towards detailed video understanding via large vision and language models.
\newblock In \emph{Proceedings of the 62nd Annual Meeting of the Association for Computational Linguistics (ACL 2024)}, 2024.

\bibitem[OpenAI(2025)]{openai2025deepresearch}
OpenAI.
\newblock Introducing deep research.
\newblock \url{https://openai.com/index/introducing-deep-research/}, 2025.
\newblock Accessed: 2025-06-05.

\bibitem[Ren et~al.(2023)Ren, Yao, Li, Sun, and Hou]{Ren2023TimeChat}
Shuhuai Ren, Linli Yao, Shicheng Li, Xu Sun, and Lu Hou.
\newblock Timechat: A time-sensitive multimodal large language model for long video understanding.
\newblock \emph{ArXiv}, abs/2312.02051, 2023.

\bibitem[Shang et~al.(2024)Shang, You, Subramanian, Darrell, and Herzig]{Shang2024TraveLERAM}
Chuyi Shang, Amos You, Sanjay Subramanian, Trevor Darrell, and Roei Herzig.
\newblock Traveler: A modular multi-lmm agent framework for video question-answering.
\newblock In \emph{Conference on Empirical Methods in Natural Language Processing}, 2024.

\bibitem[Wang et~al.(2024{\natexlab{a}})]{wang2024videoagent}
Chao-hong Wang et~al.
\newblock Videoagent: Long-form video understanding with large language model as agent.
\newblock In \emph{NeurIPS}, 2024{\natexlab{a}}.

\bibitem[Wang et~al.(2022)Wang, Li, Li, He, Huang, Zhao, Zhang, Xu, Liu, Wang, Xing, Chen, Pan, Yu, Wang, Wang, and Qiao]{Wang2022InternVideoGV}
Yi Wang, Kunchang Li, Yizhuo Li, Yinan He, Bingkun Huang, Zhiyu Zhao, Hongjie Zhang, Jilan Xu, Yi Liu, Zun Wang, Sen Xing, Guo Chen, Junting Pan, Jiashuo Yu, Yali Wang, Limin Wang, and Yu Qiao.
\newblock Internvideo: General video foundation models via generative and discriminative learning.
\newblock \emph{ArXiv}, abs/2212.03191, 2022.

\bibitem[Wang et~al.(2023)Wang, Yang, and Ren]{Wang2023LifelongMemoryLL}
Ying Wang, Yanlai Yang, and Mengye Ren.
\newblock Lifelongmemory: Leveraging llms for answering queries in long-form egocentric videos.
\newblock In \emph{arXiv}, 2023.

\bibitem[Wang et~al.(2024{\natexlab{b}})Wang, Yu, Stengel-Eskin, Yoon, Cheng, Bertasius, and Bansal]{Wang2024VideoTreeAT}
Ziyang Wang, Shoubin Yu, Elias Stengel-Eskin, Jaehong Yoon, Feng Cheng, Gedas Bertasius, and Mohit Bansal.
\newblock Videotree: Adaptive tree-based video representation for llm reasoning on long videos.
\newblock \emph{ArXiv}, abs/2405.19209, 2024{\natexlab{b}}.

\bibitem[Yang et~al.(2022)Yang, Miech, Sivic, Laptev, and Schmid]{Yang2022ZeroShotVQ}
Antoine Yang, Antoine Miech, Josef Sivic, Ivan Laptev, and Cordelia Schmid.
\newblock Zero-shot video question answering via frozen bidirectional language models.
\newblock \emph{ArXiv}, abs/2206.08155, 2022.

\bibitem[Zhang et~al.(2023)Zhang, Lu, Islam, Wang, Yu, Bansal, and Bertasius]{Zhang2023ASL}
Ce Zhang, Taixi Lu, Md~Mohaiminul Islam, Ziyang Wang, Shoubin Yu, Mohit Bansal, and Gedas Bertasius.
\newblock A simple llm framework for long-range video question-answering.
\newblock In \emph{Conference on Empirical Methods in Natural Language Processing}, 2023.

\bibitem[Zhang et~al.(2024)Zhang, Xie, Feng, Li, Liu, and Nie]{Zhang2024HCQAE}
Haoyu Zhang, Yuquan Xie, Yisen Feng, Zaijing Li, Meng Liu, and Liqiang Nie.
\newblock Hcqa @ ego4d egoschema challenge 2024.
\newblock \emph{ArXiv}, abs/2406.15771, 2024.

\bibitem[Zhi et~al.(2025)Zhi, Wu, shen, Li, Li, Shao, and Zhou]{Zhi2025VideoAgent2ET}
Zhuo Zhi, Qiangqiang Wu, Minghe shen, Wenbo Li, Yinchuan Li, Kun Shao, and Kaiwen Zhou.
\newblock Videoagent2: Enhancing the llm-based agent system for long-form video understanding by uncertainty-aware cot.
\newblock \emph{ArXiv}, abs/2504.04471, 2025.

\end{thebibliography}
